\documentclass[11pt]{article}
\usepackage[margin=2cm]{geometry}
\usepackage[authoryear]{natbib}
\usepackage{graphicx}
\usepackage{amsmath}
\usepackage{authblk}
\usepackage[utf8]{inputenc} 
\usepackage{float}

\graphicspath{figures/}

\title{Lazy learning: a biologically-inspired plasticity rule for fast and energy efficient synaptic plasticity.}
\author[1]{Aaron Pache \thanks{lpxap7@nottingham.ac.uk}}
\author[1,2]{Mark CW van Rossum\thanks{mark.vanrossum@nottingham.ac.uk}}
\affil[1]{School of Mathematical Sciences}
\affil[2]{School of Psychology, University of Nottingham, Nottingham, UK}

\begin{document}
\maketitle
\begin{abstract}
When training neural networks for classification tasks with backpropagation, parameters are updated on every trial, even if the sample is classified correctly. In contrast, humans concentrate their learning effort on errors.
Inspired by human learning, we introduce lazy learning, which only learns on incorrect samples. Lazy learning can be implemented in a few lines of code and requires no hyperparameter tuning. Lazy learning achieves state-of-the-art performance and is particularly suited when datasets are large. For instance, it reaches 99.2\% test accuracy on Extended MNIST using a single-layer MLP, and does so 7.6$\times$ faster than a matched backprop network.
\end{abstract}

Recent progress in machine learning has been partly attributed to the use
of large data sets \citep{LeCun2015}. Even already large datasets are often augmented to further boost performance. However, repeatedly cycling
over large datasets and adjusting the parameters
is time and energy consuming. In classification tasks, backprop
typically prescribes synaptic updates regardless of whether the classification
was correct or incorrect; updating the network to be correct if it
was wrong, but also updating to be \textit{more} correct if it was right. Is this incessant updating necessary or do more efficient training rules
exist?

Previous work has determined that high accuracies for many datasets can be achieved using only a small number of critical samples, called the coreset \citep{Agarwal2005coreset}. Methods have been developed to find these samples and focus training \citep{Toneva2019unforgettable,paul2021deepdiet,coleman2020selectionproxy}. These methods require a model trained beforehand to identify key samples and form a coreset dataset.
However, for a new dataset, it won't be clear what the coreset is or even whether the coreset is small or large. A simple algorithm that restricts learning and does not need tuning is lacking.



Humans appear to be more efficient by placing far more importance
on mistakes and errors, a phenomenon known
as the negativity bias \citep{Rozin2001}. A familiar example is the
sharp `jolt' we experience when typing a word incorrectly. When such
mistakes occur a large event-related potential known as the error-related
negativity (ERN) is elicited in the electroencephalogram signal (EEG),
which usually precedes behavioural changes like post-error slowing
and post-error improvement in accuracy \citep{Kalfaoglu2017}. The
ERN is linked to dopamine \citep{Holroyd2002} - a 
plasticity modulator required for persistent forms of plasticity \citep{OCarroll2004}. 
Indeed, subjects learn better on stimuli that evoke a larger EEG \citep{debruijn2020}.

This suggests that, in contrast to backprop, biological learning largely occurs on mistakes, while skipping
over correct samples. The reason might be that learning is emerging as a metabolically costly process. For instance, 
flies subjected to  aversive conditioning  and subsequently
starved, died 20\% sooner than control flies \citep{Mery2005b}.
Similarly, flies doubled sucrose consumption after long term memory formation \citep{Placais17,Placais13}. While the
reason for these costs are unknown, they were recently estimated to
be as much as 10mJ/bit \citep{Girard2023}.

The high energetic cost of learning is not unique to biology and is
also important in traditional computers and in ANNs. Therefore restricting
synaptic updates to incorrect samples could also benefit the training
time and energy requirements of computer implementations of neural
networks. 

\section*{Results}

With commonly used loss functions backprop updates the network parameters even if an example is classified correctly. For instance when using cross-entropy loss and one-hot encoding, weights are updated unless the network output exactly equals the target one-hot vector. In contrast, the idea in lazy learning is to only update the weights when a sample is misclassified; not unlike the classic perceptron algorithm \citep{Rosenblatt1958}. We describe two variants: `pure lazy learning' and `lazy learning'. In \textbf{pure lazy learning}, the network \textit{only} learns when the presented sample is incorrectly classified; if the sample is correctly classified, no update occurs and the algorithm skips to the next example. This amounts to a single line of code
\begin{equation}
    \texttt{if (pred != target): \#do backprop}.
\end{equation}
As is shown below, pure lazy learning can lead to fragile generalization performance. We therefore also introduce
\textbf{lazy learning} which instead remembers all incorrect samples and learns when the sample is currently misclassified or was at \textit{any previous time} misclassified. Lazy learning maintains a vector of booleans with as many indices as there are samples in the dataset and is initialized to zero. If at any time during learning a sample is misclassified, the index of that sample in the vector is set to 1, which signifies remembering that sample for the rest of the run. The learning rule is 
\begin{equation}
\texttt{if remember[index] == 1: \#do backprop}
\end{equation}
Neither rule requires hyperparameter tuning and both are very easily added to existing code.  

\subsection*{Intuition from a simple task}

\begin{figure}
    \centering
    \includegraphics[width=0.8\linewidth]{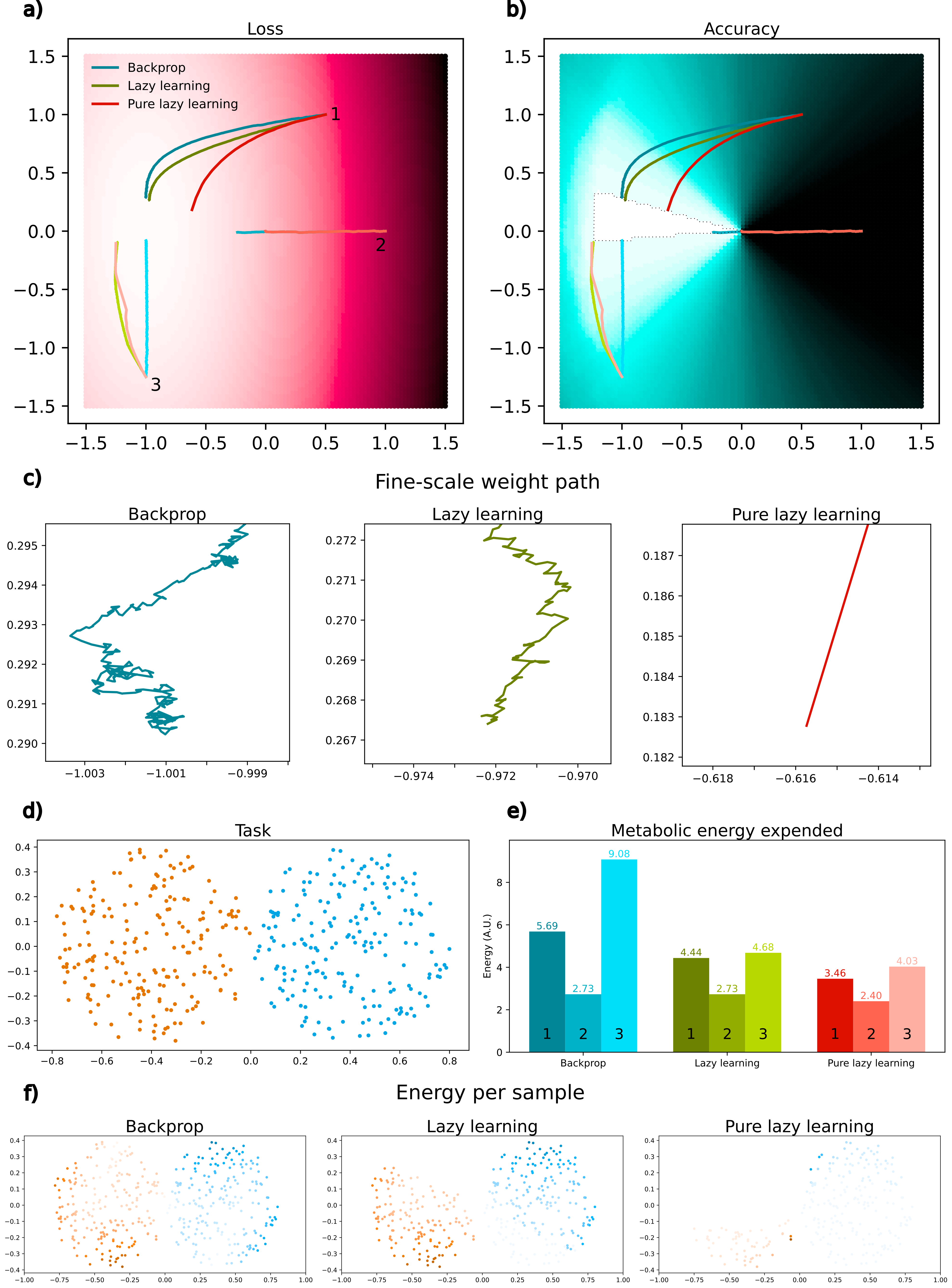}
    \caption{Lazy learning on a two-class problem in two dimensions (panel d).
    \textbf{a, b}. The loss surface (a) and accuracy surface (b) as function of the two synaptic weights, with the weight trajectories of 3 initial conditions for each learning rule: backprop (blue), lazy learning (green) and pure lazy learning (red). On a coarse scale backprop  follows the gradient of the loss. Pure lazy learning instead follows the  gradient of the accuracy surface. Lazy learning displays both backprop like characteristics, tracing backprop's path exactly from initial condition 2, as well as pure lazy learning's ability to trace the accuracy surface in initial condition 3. \textbf{c}. Zoom on the final part of the weight path of each learning rule for initial condition 1. \textbf{d}. Distribution of the inputs. The two classes are orange and blue. \textbf{e}. The energy consumed across initial conditions and learning rules. Regardless of initial conditions, pure lazy learning consumes the least energy. \textbf{f}. The energy expended on each sample for initial condition 1. Darker colors correspond to more energy.}
    \label{fig:simple_task}
\end{figure}

We first illustrate the lazy learning rules with a simple, linearly separable classification task of two data clouds using a small learning rate (0.001), Fig.\ref{fig:simple_task}. Given our biological perspective, we make no use of data normalization, data augmentation, momentum, optimizers or batching as these would raise questions in terms of the biological implementation. While more work is needed, preliminary investigations indicate that such modifications do not negate lazy learning's benefits.

We track the weight evolution starting from 3 different initial conditions on both the loss and accuracy surface, Fig.\ref{fig:simple_task}a+b. Starting from initial condition 1, on a coarse level standard backprop follows a typical graded descent trajectory towards the minimum of the continuous loss surface. At a finer scale backprop updates irrespective of whether the sample is correctly classified, in the direction of that sample's gradient, resulting in a windy path, Fig.\ref{fig:simple_task}c. Its weight trajectory resembles a random walk with drift.

In contrast, pure lazy learning follows a path that on a coarse level perpendicularly ascends the terraced accuracy surface. It stops as soon as it reaches the 100\% accuracy terrace (white, delineated region in Fig.\ref{fig:simple_task}b). At a finer-scale, the weight trajectory of pure lazy learning gets progressively straighter during training as fewer samples contribute to learning. At the end of learning a single sample dictates the direction, resulting in a straight path Fig.\ref{fig:simple_task}c. Lazy learning follows an intermediate trajectory. Trajectories from other initial conditions are consistent with this view. 
Initial condition 3 highlights that in contrast to backprop,  lazy learning ascends the accuracy surface.

In lazy learning updates are less frequent, which translates to faster training time and fewer updates in computers (see Appendix). In light of our biological motivation, we would also like to know the metabolic energy savings achieved by lazy learning. As the details of biological energy expenditure associated with learning are not well known, we use a previously proposed
energy measure \citep{li2020energy},
\begin{equation}
    M=\sum_{i=1}^{N}\sum_{t=1}^{T}|w_{i}(t)-w_{i}(t-1)|\label{eq:energy}
\end{equation}
where $N$ is equal to the total number of weights in the network
and $T$ is equal to the total number of learning steps. The metabolic energy equals the L1-norm of the weight trajectory.

Fig.\ref{fig:simple_task}e compares this energy measure across the learning rules; pure lazy learning is most efficient, but lazy learning is almost as efficient. The amount of saving depends on the initial condition. For instance, from initial condition 2, the paths of all rules are identical on a coarse level and little saving can be obtained, aside from pure lazy learning stopping earlier.

To further explore the origin of the savings, Fig.\ref{fig:simple_task}f colors the data such that darker color means that more energy was expended on that sample. In standard backprop the coloring is fairly uniform (there are still variations in the weight changes that are reflected in the energy). In contrast, pure lazy learning automatically concentrates on a few edge samples - similar to Support Vector Machines. For lazy learning, the separating line corresponding to initial condition 1, which rotates   anti-clockwise during learning, is visible.

Initial condition 2 in particular highlights that pure lazy learning finds a fragile solution, where a small perturbation rightwards would lead to a large change in accuracy. This reduces the generalization performance of pure lazy learning, it is however rescued by the memory aspect of lazy learning.

\subsection*{Learning from the memory of mistakes: lazy learning of MNIST}

Having shown the principle of lazy learning on simple tasks, we apply it the MNIST task. We train MLPs with a single hidden layer, implemented in the PyTorch framework
\citep{Paszke2019pytorch}, initialized using the PyTorch
default for linear layers (kaiming uniform) \citep{He2015uniform}.
Networks contain 100, 200, 500, or 1000 hidden units and
use the Leaky Rectified Linear Unit activation function with the default
PyTorch negative slope parameter (0.01). The target is a one-hot vector. As cross entropy loss minimization requires more energy (not shown), we use a MSE cost function with a learning rate of 0.01. For biological realism we use stochastic gradient-descent, presenting a sample one at a time.
We found that {\em pure} lazy learning quickly gives good training set performance, but finds fragile solutions that don't generalize as well as backprop (see Appendix). 

\begin{figure}
    \centering \includegraphics[width=1\linewidth]{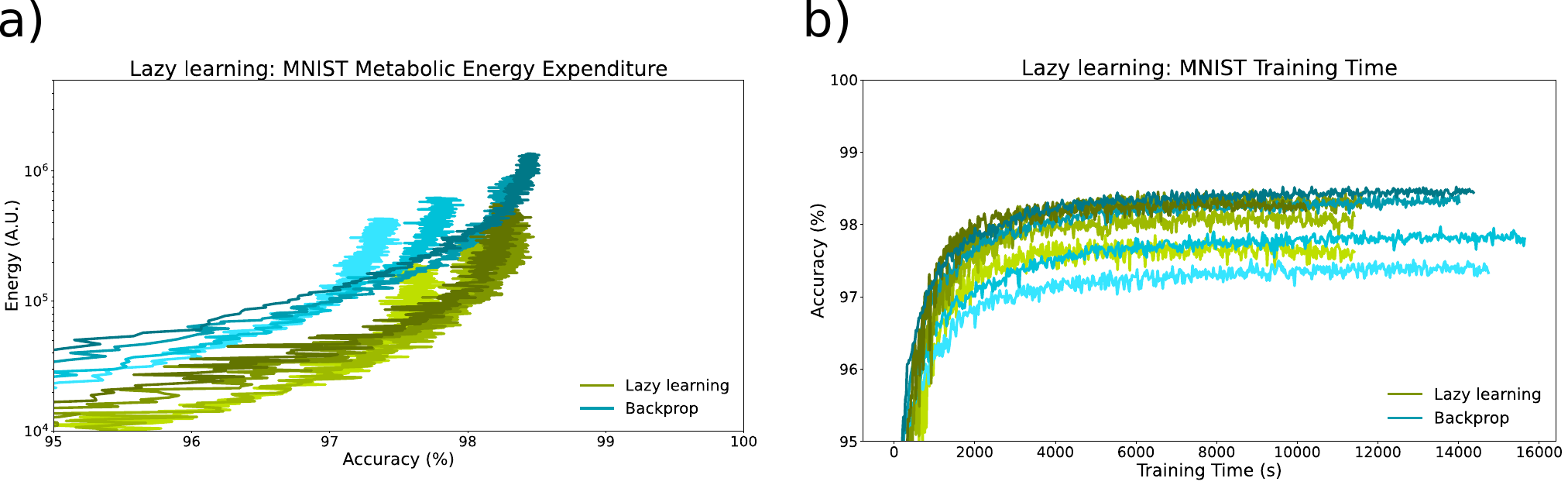}
    \caption{Lazy learning in the MNIST task.
    \textbf{a} - The metabolic energy expenditure of lazy learning for network sizes 100 (light), 200, 500, 1000 (dark) hidden units. Lazy learning saves energy while maintaining the generalization
    performance of backprop. \textbf{b} - Training times of lazy and
    backprop networks. In particular in small networks, lazy learning reaches similar performance in less time.}
    \label{fig:ll_mnist} 
\end{figure}

Instead, lazy learning learns over all samples currently or previously misclassified and should overcome the shortcomings of pure lazy learning. To test this we track both the energy expended and testing accuracy as training proceeds and plot them against each other, Fig.\ref{fig:ll_mnist}a. 
Reaching higher accuracy requires more energy and larger networks can reach better performance.  Importantly, lazy learning always expends substantially less energy (note the logarithmic y-scale). Also the total number of parameter updates was always less (not shown).

To characterize the speed of lazy learning, we can not use the number of epochs. Compute time is dominated by parameter updating rather than by inference. While lazy learning requires the entire training dataset for the forward pass, it only uses previously misclassified samples for the backward pass. Thus while lazy learning typically takes more epochs, each epoch takes much less time because it skips updating for most samples.
Instead we measure the computing time spent, Fig.\ref{fig:ll_mnist}b (see Methods for details).
The training times only include the time to perform feedforward and backward updates and exclude the fixed (small) overhead time for testing accuracies and losses. Training times are typically faster for lazy learning, while reaching similar accuracy levels, Fig.\ref{fig:ll_mnist}b. Thus by memorizing the critical samples, lazy learning reduces training time by finding more robust solutions than pure lazy learning and maintains similar energy savings.

\subsection*{Lazy learning of Extended MNIST}

We hypothesized that with more data, lazy learning could perform even
better since the directions it takes on the loss surface are dictated
by misclassified samples. In larger datasets there are more informative
samples to the true direction of a 100\% accuracy terrace. To test
this, we repeat the experiment on the EMNIST digits task which is identical to MNIST, but it contains 240,000 training samples and 40,000 testing samples (4$\times$ the size of MNIST).

\begin{figure}[H]
    \centering \includegraphics[width=1\linewidth]{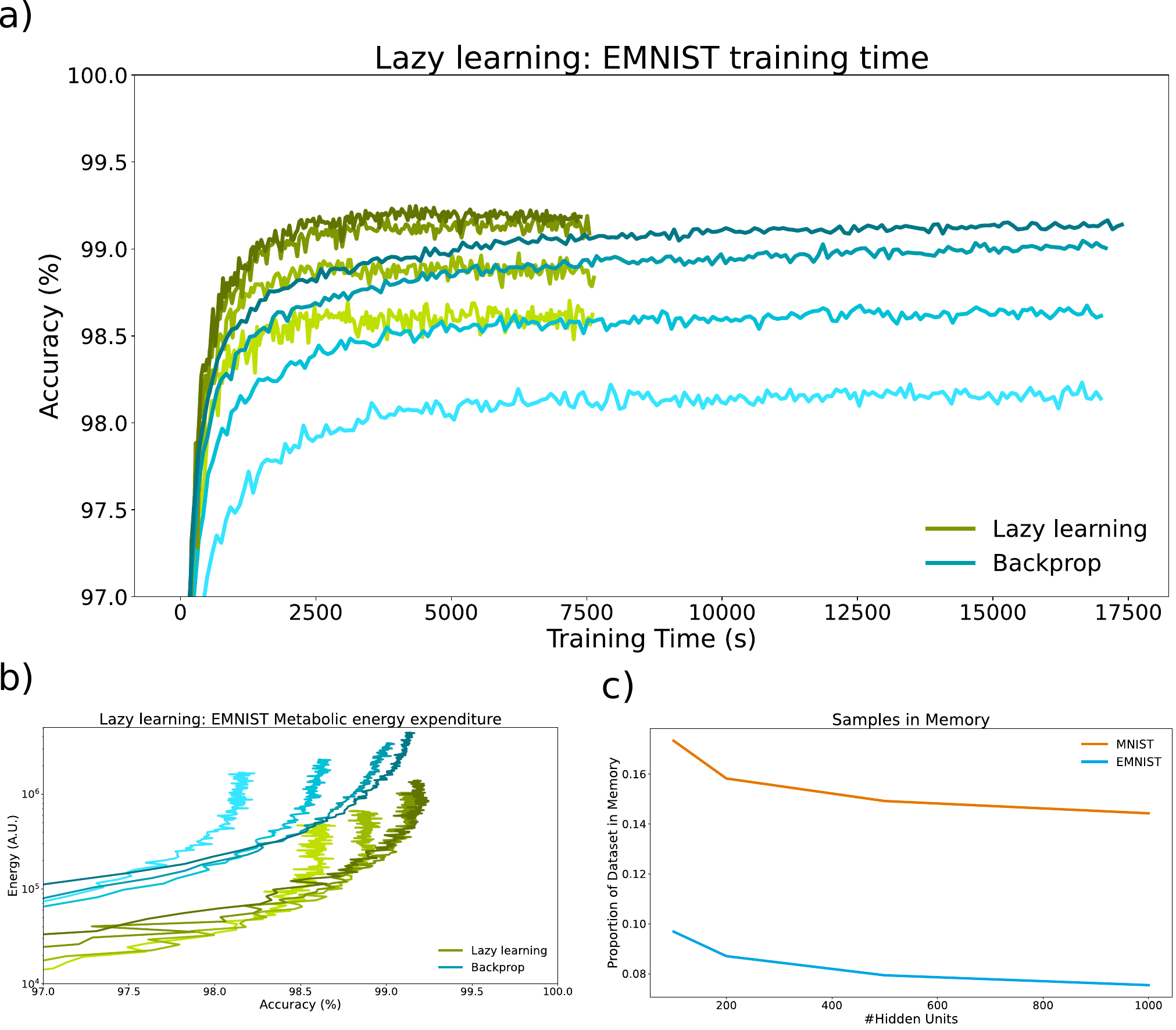}
    \caption{Lazy learning in the extended MNIST task.
    \textbf{a} - The training times of lazy and
    backprop networks for sizes 100 (light), 200, 500, 1000 (dark) hidden units. To train over the 50 epochs, lazy learning takes
    less time than backprop networks because they perform synaptic updates
    on less samples. Note: this training time excludes time taken to pass over the test set since this is a fixed overhead. \textbf{b} - The metabolic energy expenditure of lazy learning for
    network sizes (100, 200, 500, 1000), darker colors indicate larger
    networks. Lazy learning is able to both save energy and maintain the
    performance of backprop. \textbf{c} - The proportion of samples memorized
    by lazy learning. When trained on MNIST they memorize at most
    17\% of the dataset, when trained on EMNIST, they memorize at most 10\% of
    the dataset. This represents the source of the training time improvements.}
    \label{fig:ll_emnist} 
\end{figure}

Compared to backprop, lazy learning improves accuracy over backprop, particularly in smaller networks, Fig.\ref{fig:ll_emnist}a. It has been suggested that as network size increases, the number of local minima increases and further these minimas are no worse than
the global minima \citep{Kawaguchi2018many}. We conjecture this is
because the loss surface approximates the accuracy surface and the loss surfaces of larger networks are better able to approximate
the accuracy surface. However, small lazy networks can approximate the accuracy surface better than small backprop networks.

Moreover, lazy learning also significantly improves the training time across all network sizes. In particular, the 1000 unit backprop network achieves a maximum testing accuracy of 99.165\% within the 50 epochs, the lazy network reaches the same accuracy 7.6$\times$ faster. When tasked on EMINST, lazy learning saves even more energy in comparison to the savings of MNIST, Fig.\ref{fig:ll_emnist}b. The more samples in a dataset, the more random of a walk backprop takes in weight space and therefore the more energy it consumes. 

Lazy learning saves proportionally more energy by training on proportionally less samples in EMNIST, compared to MNIST, Fig.\ref{fig:ll_emnist}c.
Interestingly, in absolute numbers lazy learning memorizes twice as much samples from a four times larger dataset. It uses these extra samples to improve performance over MNIST. The training time speed up is roughly proportional to the fraction of memorized samples and the relative gain in generalization these samples provide. It would be of interest to understand the general relation between the dataset size, task complexity, and the number of samples used by lazy learning.



\section*{Discussion}

Learning is a costly process, both for biology and computer hardware. To reduce this cost lazy learning draws inspiration from the observation that humans pay more attention to errors when learning.
For AI practitioners, lazy learning can lead to faster, energy efficient and more performant synaptic plasticity rules. Code complexity and memory overhead are minimal. We foresee various applications.
Lazy learning acts as a ‘plug-and-play’ method of self-paced curriculum learning \citep{Bengio2009curriculum}. As the network improves, misclassified samples will lie closer to decision boundaries or be an anomalous representation of the class. It has been shown that a dataset can be reduced to the number of repeatedly misclassified samples \citep{Toneva2019unforgettable}. Similarly, the set of samples lazy learning uses can be used as a coreset dataset. Here, the loss minima of these samples correspond to the accuracy terrace solution found by the lazy network.

Since lazy learning only learns on mistakes it may also be useful for class imbalanced datasets \citep{Johnson2019imbalancereview}, since imbalanced classes could be classified incorrectly more often. In this case, lazy learning would naturally balance the dataset. Lazy learning also regularizes the weights which could improve generalization performance. 

 Lazy learning could serve a unique purpose in Difference Target Propagation \citep{Lee2014dtp} and Direct-Difference Target Propagation \citep{Meulemans2020ddtp}. These have been proposed as a biological implementation of backprop, with separate forward and backward weights. The backward weights attempt to recreate the hidden activities achieved by the forward weights, effectively inverting the higher unit activities to the lower unit activities. Here, lazy learning could also signal separate forward and backward synaptic updates. When the network is incorrect, both forward and backward updates occur. However, when the network is correct, only backward updates occur, which may help stabilize the backward inversions.

 Recently, there is evidence that Transformer networks store the weights of a smaller neural network within them and perform gradient-descent from a sequence of presented examples in the input without updating its own weights \citep{Akyurek2022transformer}. Furthermore, it has been proposed that the hippocampus of the brain mimics the representations found by that of a transformer \citep{Whittington2022hippocampus}. Lazy learning effectively memorizes a small subset of a much larger dataset through an attention mechanism. Taken together with the results presented here, this suggests the brain may learn from memorized examples within the hippocampus and that this is energy efficient.

 Lazy learning is also of importance for neuroscientists. The hippocampus has long been suggested as a temporary buffer to allow for offline learning. Lazy learning suggests that the content of the hippocampus is carefully titrated. On one hand, only events that evoke an error signal need to be stored. On the other hand, the result of pure lazy learning show that such events should be kept in memory, even after they do not evoke an error anymore. 
 
Though neural networks train on millions of examples within a dataset, we show this training can be reduced with an attentional mechanism that focusses misclassified samples.

\section*{Methods}

For the simple task, a simple, single-layer perceptron consisting
of 2 weights, $w_{i}$, is used. It takes two inputs, $x_{i}$, that
represent the cartesian vector of a data point. The task is to determine
a linearly separable binary decision, between two circles as shown
in Fig.\ref{fig:simple_task}d. 
We used a MSE loss function between target and $h=\mathbf{w}\cdot\mathbf{x}$, and classify using $y=H(h)$, where $H()$ is the Heaviside function.

The networks are trained on the MNIST dataset
\citep{Lecun1998mnist} which consists of 28$\times$28 grey-scale
pixel images of handwritten digits 0 - 9. There are 60,000 training
samples which are presented in a random order and 10,000 testing images.
Every 5,000 training samples, various metrics including loss, accuracy
and energy are measured on the test and training set. All lazy learning simulations
ran for 50 epochs. 

For timing purposes, all compared simulations of backprop and lazy learning networks run on the same machine. This was either an Nvidia RTX A6000, Threadripper 3970X, 128GB of RAM or an Nvidia GTX 1080Ti, Ryzen 1800X and 32GB of RAM. Times were obtained using the \texttt{timeit} python library.
In the case of lazy learning, the MNIST simulations ran using the Nvidia GTX 1080Ti setup on the GPU. 
We also use the EMNIST digits dataset \citep{Cohen2017emnist}, which used the Nvidia RTX A6000 setup on the GPU and measured metrics every 60,000 samples. Depending on the hardware and various hyperparameter setups, different training times may be obtained. If all samples are misclassified then lazy learning will approach an identical solution to backprop. Generally lazy learning is faster than backprop but will otherwise be just as fast.

For pure lazy learning on MNIST (see Appendix) we set a cut-off accuracy of 97.5\% or alternatively, gave up at 50 epochs. These simulations used the Nvidia A6000 setup on the GPU. 


\subsection*{Acknowledgments}

It is a pleasure to thank Claudia Danielmeier and Josh Khoo for discussion.
This research was supported by a grant from NVIDIA and some simulations utilized an
NVIDIA RTX A6000. 

\section*{Appendix}
\subsection*{Pure lazy learning in complex tasks}
Here we show the results of pure lazy learning on the MNIST task.

In addition to the energy listed in Eq. \ref{eq:energy}, we also define a minimal energy. The lowest energy solution would follow a straight path in weight space to the solution closest to the initial condition. However, since finding the optimal solution is intractable, we can designate a candidate ideal solution to be the L1-norm of the weights at a given point in time. This would assume the network traced a straight line to a given level of performance.
\begin{equation}
    \centering
    M_{\textrm{min}} = \sum_{i=1}^{N}|w_{i}(T)-w_{i}(0)|
    \label{eq:minimum}
\end{equation}
While similar to Eq. \ref{eq:energy}, it only uses the initial and final weights. Eq. \ref{eq:energy} and Eq. \ref{eq:minimum} define an energetic inefficiency, 
\begin{equation}
    \centering
    \textrm{Ineff} = \frac{M}{M_{\textrm{min}}}
    \label{eq:inefficiency}
\end{equation}
Conceptually, it describes how tortuous the weight path is compared to the straight path. Consequently there are two methods to save energy: solutions that have a smaller norm (relative to the initial conditions) and less windy paths.

\begin{figure}
    \centering 
    \includegraphics[width=0.8\linewidth]{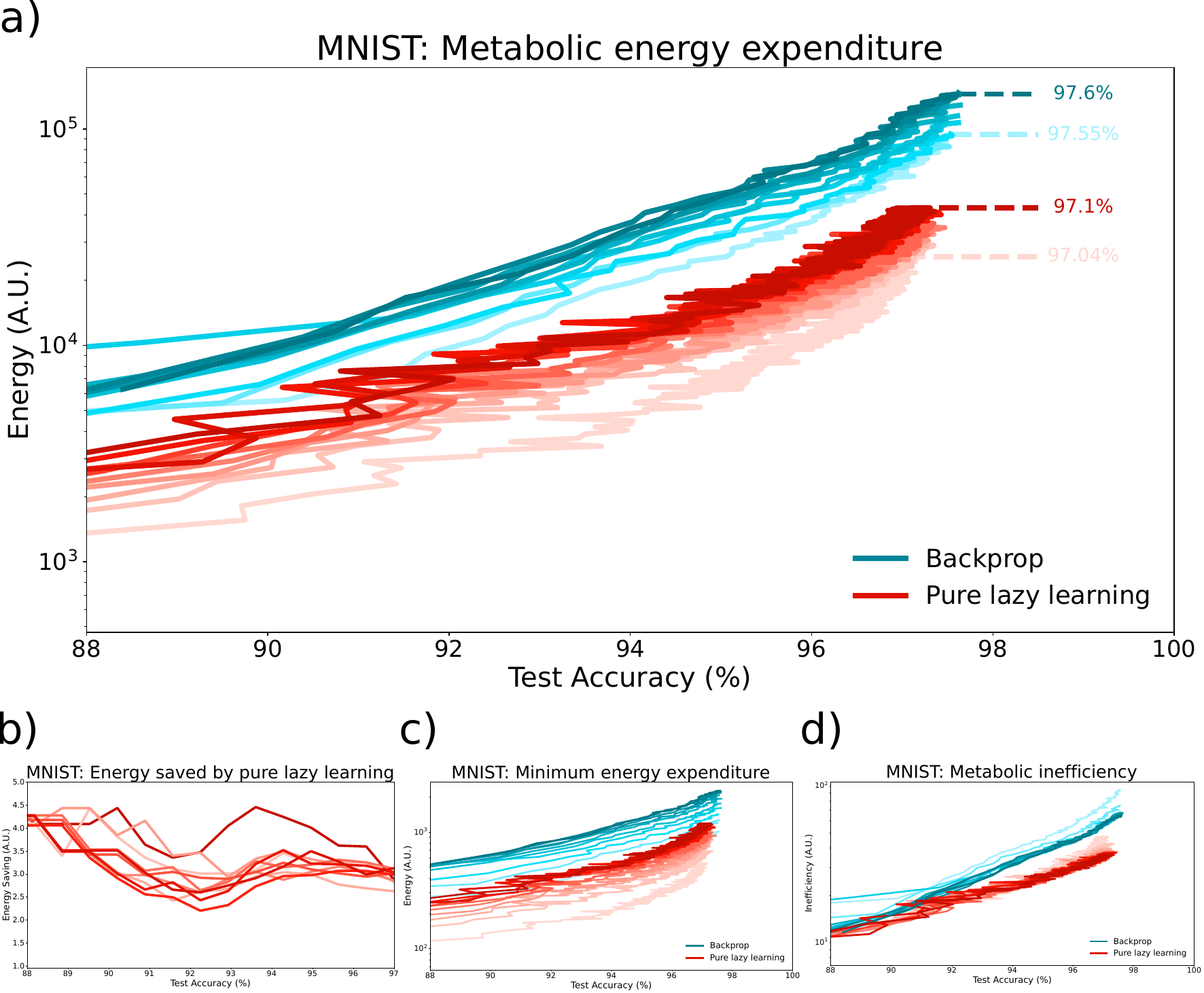}
    \caption{Pure lazy learning in the MNIST task. \textbf{a} - The metabolic energy consumed by backprop and lazy learning networks for network sizes 100, 200, ..., 1000. Increasing darkness represents increasing network size. The testing accuracy
    reached by the end of training for the smallest and largest networks
    under each learning regime are indicated. Training stops after 50 epochs or reaching
    97.5\% testing accuracy. For all network sizes and for all accuracies
    achieved by lazy learning, lazy learning saves considerable metabolic
    energy over backprop but is unable to achieve the same performance.
    \textbf{b} - The energy saved by lazy learning for all achieved accuracies.
    Lazy learning can save between 2.2-4.5x of the metabolic energy expended
    by backprop. \textbf{c} - Lazy learning achieves a lower minimum
    energy in comparison to the equivalent  backprop networks,
    indicating that lazy learning networks find a shorter weight path
    to attain a given accuracy.  \textbf{d} - The energy inefficiency
    of backprop  and lazy learning  networks. Lazy learning
    is more efficient than backprop, indicating that lazy learning follows
    a less windy path to a given accuracy. }
    \label{fig:pll_mnist}
\end{figure}

Pure lazy learning saves significant energy over backprop
for all network sizes and for all accuracies attained by pure lazy
learning, Fig.\ref{fig:pll_mnist}. The energy savings fall between 2.2-4.5x that of backprop and this variance is likely a result of the potentially fragile
solutions pure lazy learning finds, as demonstrated in Fig.\ref{fig:simple_task}.
Since the minimum energy of pure lazy learning is less than backprop, the
final L1-norm of the weights is smaller and therefore saves energy, Fig.\ref{fig:pll_mnist}c. It also shows that pure lazy learning has a regularizing effect on the weights.

The inefficiency of lazy learning is also less than that of backprop, indicating the weights follow a less tortuous path, Fig.\ref{fig:pll_mnist}d. Pure lazy learning thus saves energy by following a less tortuous trajectory to a closer solution.

\begin{figure}
\centering \includegraphics[width=1\linewidth]{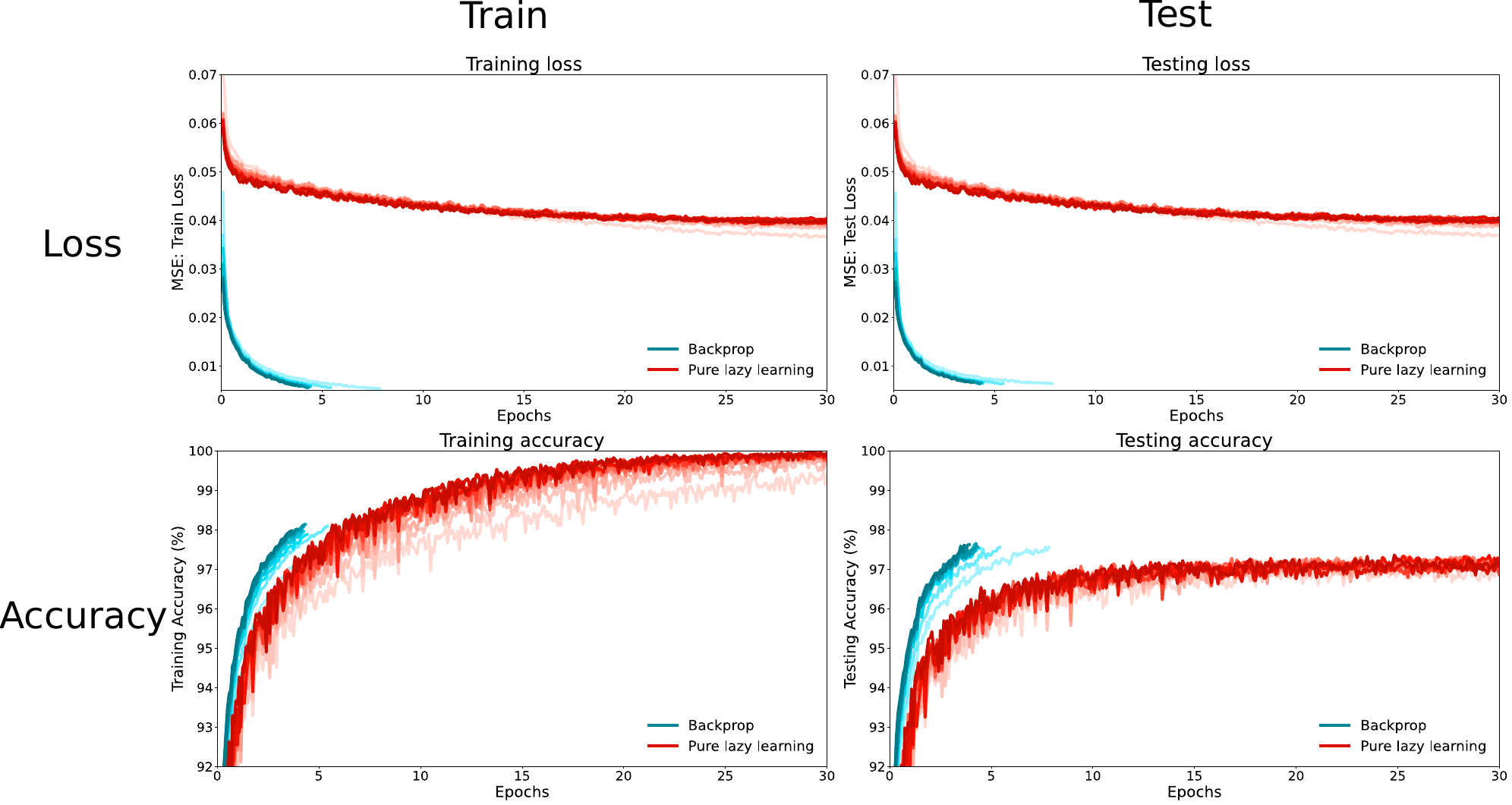}
\caption{The test and training, loss and accuracy metrics of pure lazy learning for network sizes 100, 200, ..., 1000.
Pure lazy learning has a significantly higher loss than backprop in
both testing and training (top). This stems from its ability to allow
low confidence but correct classifications. Also significant is lazy
learning's training accuracy, which continues to improve towards 100\%,
but the testing accuracy levels off. Since the testing accuracy levels
off, as opposed to dropping, this suggests lazy learning is not necessarily
overfitting. In all cases, lazy learning is unable to achieve the
threshold testing accuracy of 97.5\%, while equivalent backprop networks
reach this within 10 epochs. This performance marks a deficit to pure
lazy learning.}
\label{fig:pll_metrics} 
\end{figure}

One shortcoming of pure lazy Learning is the maximal performance it
can reach. While backprop networks were able to achieve the 97.5\%
threshold testing accuracy within 50 epochs, pure lazy learning networks
were not. We highlight this performance in Fig.\ref{fig:pll_metrics},
which depicts the test and training loss alongside the test and training
accuracy for both backprop and pure lazy learning (up to 30 epochs).
Significantly, pure lazy learning reach a loss 10-fold larger than
that of backprop networks. This occurs since pure lazy learning can
tolerate low confidence but correct neuron outputs to a given sample.
They end up finding discrete, sharp terraces where tiny weight perturbations
can cause large differences in accuracy.

The loss demonstrates that for MNIST, some 95\% accuracy terraces
lie around a mean-squared error loss of 0.07 and is significantly
higher than the loss backprop requires to reach the same accuracy.
This evinces how high some of the 95\% terraces might be on the loss
surface. 

While the simple example shown in Fig.\ref{fig:simple_task}
provides intuition to distinguish between the terraced accuracy surface
and the loss surface. In more complicated tasks, there are many different samples with different varying tolerances for their accuracy terrace and this results
in many sharp fragile solutions at the intersection of these terraces. Regardless, in a more complex task, pure lazy learning is able to proficiently locate high test accuracy terraces using less energy in comparison to backprop. 

The more correct pure lazy learning is, the less samples are evaluated
in an epoch. One incorrectly labelled sample could dominate the gradients
prescribed by gradient-descent and under the pure lazy learning rule
this moves pure lazy learning outside of better test accuracy terrace.
For example, index 59915 of MNIST labels a `7' looking digit as a
`4' and would guide lazy learning towards a fragile accuracy terrace representative of the training set accuracy terrace
instead of the global terrace.

Though pure lazy learning is unable to achieve the 97.5\% testing
accuracy, the training accuracy continues to improve. It is important
to note the test accuracy does not begin to fall which would indicate
overfitting. We infer that pure lazy learning continues to seek the 100\%  accuracy terrace of the training set and this remains on the outskirts of the
test set when seeking minimal energy consumption from a particular
initial condition.


\subsection*{Number of synaptic updates and CPU time}
The energy measure used in the main results is useful as biological model and to indicate the tortuousness of a weight path. However, it is not as useful for computational efficiency. An alternative energy measure is the number of synaptic updates. Under this energy measure, lazy learning also saves energy; even more than using the biological measure used in the main text. 

In the main results, networks are trained using GPUs. To show the generality of the result we here use the CPU time taken. On a CPU, small networks train faster, but large networks take significantly longer to train. 


\begin{figure}[H]
    \centering \includegraphics[width=1\linewidth]{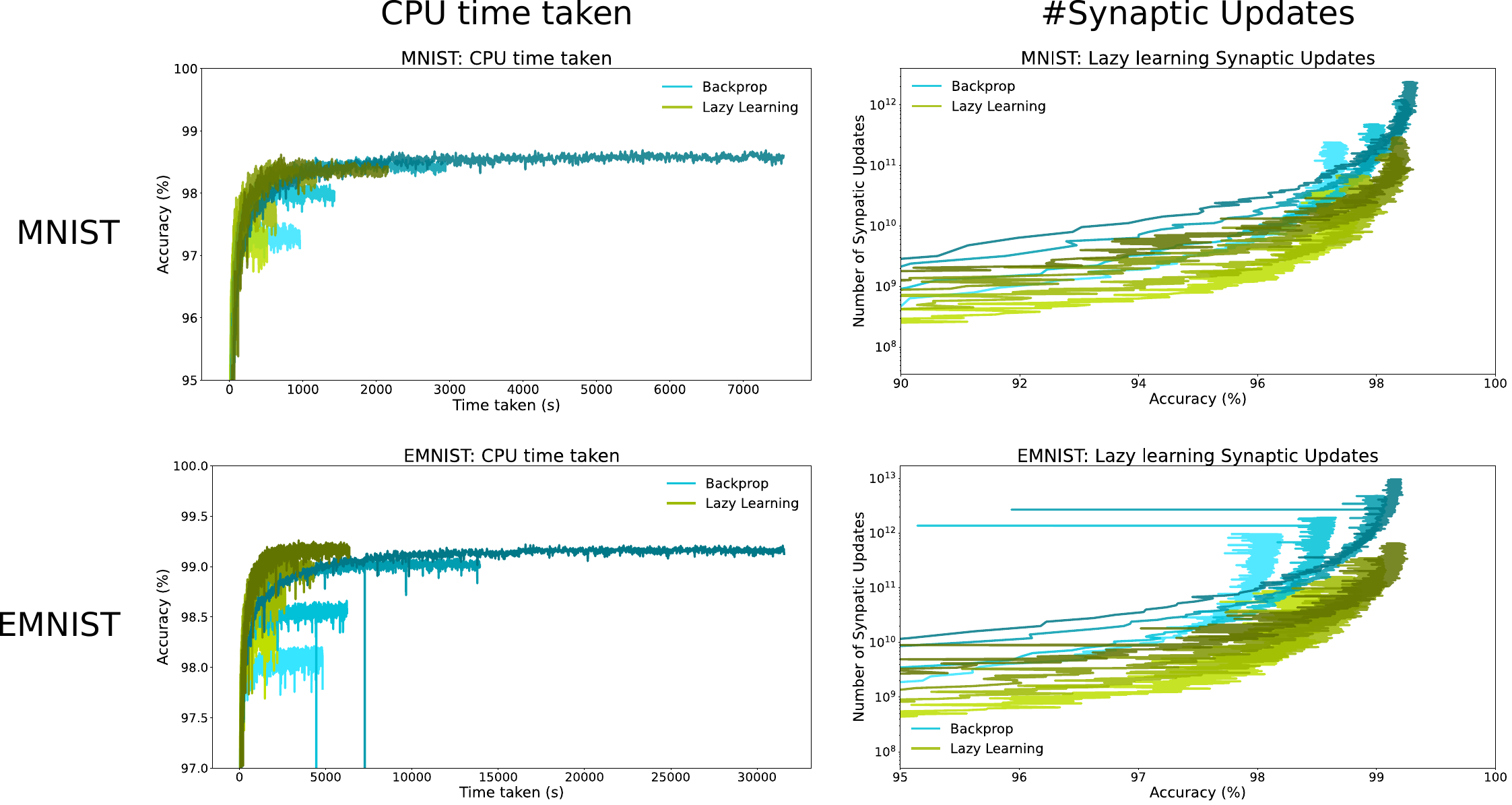}
    \caption{The CPU time taken and number of synaptic updates energy of backprop and lazy learning for network sizes (100, 200, 500, 1000). On a CPU, larger networks take longer to train over the 50 epochs. Since lazy networks update less often, it takes significantly less time to train. Lazy learning uses around 10-fold less synaptic updates than backprop to reach an equivalent accuracy.}
    \label{fig:cpu_and_m0} 
\end{figure}

\bibliographystyle{abbrvnat}
\bibliography{references.bib}

\end{document}